\documentclass[letterpaper, 10 pt, conference]{ieeeconf}  

\IEEEoverridecommandlockouts                              

\usepackage{amsmath,amssymb,amsfonts}
\usepackage{algorithmic}
\usepackage{graphicx}
\usepackage{textcomp}
\usepackage{xcolor}
\usepackage{comment}
\usepackage[table,xcdraw]{xcolor} 
\usepackage{makecell}
\newcommand{\mc}[1]{\multicolumn{1}{|l|}{#1}}
\usepackage{caption}

\usepackage{hyperref} 
\usepackage[
backend=biber,
style=numeric,
sorting=none
]{biblatex}
\newcommand{\ourmodel}{\textsc{Ventura}}

\newcommand{\seclabel}[1]{\label{sec:#1}}
\newcommand{\secref}[1]{Sec.~\ref{sec:#1}}
\newcommand{\sseclabel}[1]{\label{ssec:#1}}

\newcommand{\figlabel}[1]{\label{fig:#1}}
\newcommand{\figref}[1]{Fig.~\ref{fig:#1}}
\newcommand{\tablabel}[1]{\label{tab:#1}}
\newcommand{\tabref}[1]{Table~\ref{tab:#1}}

\newcommand{\akz}[1]{[{\textbf{\textcolor{magenta}{Arthur: #1}}}]}

\begin{document}

\title{\LARGE \bf
\ourmodel{}: Adapting Image Diffusion Models for Unified Task Conditioned Navigation 
}
\author{Anonymous Author(s)}
\definecolor{navy}{RGB}{0,0,128}

\author{%
  Arthur Zhang$^{1,2}$\textsuperscript{*},
  Xiangyun Meng$^{2}$\textsuperscript{\ddag},
  Luca Calliari$^{2}$\textsuperscript{\dag},
  Dong\mbox{-}Ki Kim$^{2}$\textsuperscript{\ddag},
  \\
  Shayegan Omidshafiei$^{2}$\textsuperscript{\ddag},
  Joydeep Biswas$^{1}$\textsuperscript{*},
  Ali Agha$^{2}$\textsuperscript{\ddag},
  Amirreza Shaban$^{2}$\textsuperscript{\ddag}
  \\[0.3em]
  $^{1}$UT Austin \quad $^{2}$Field AI
}

\twocolumn[{%
\renewcommand\twocolumn[1][]{#1}%
\maketitle
\begin{center}
    \vspace{-20pt}
    \centering
    \includegraphics[width=1.0\textwidth]{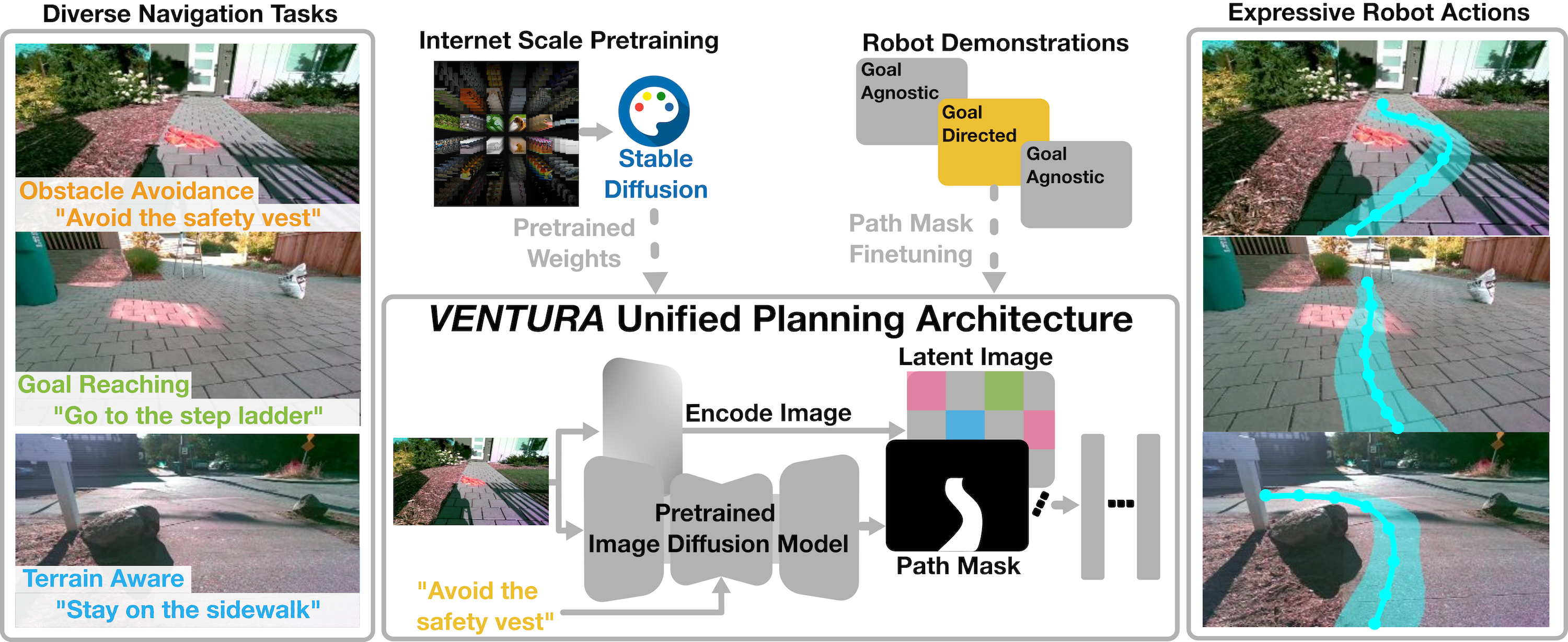}
    \captionof{figure}{Given an image and a language instruction (e.g. \emph{avoid the safety vest}), \ourmodel{} uses a fine-tuned image diffusion model to render a path mask in the image space. The path mask is then passed to a lightweight policy network to produce executable robot actions. By training on a mix of goal-agnostic and goal-directed demonstrations, \ourmodel{} grounds diverse language instructions in safe and precise robot motions.}
    \label{fig:header}
    \figlabel{mainapproach}
\end{center}%
}]

\IEEEpeerreviewmaketitle

\begingroup
\renewcommand\thefootnote{\fnsymbol{footnote}}
\footnotetext[1]{\texttt{\{ arthurz, joydeepb \} @cs.utexas.edu}}
\footnotetext[2]{\texttt{lcallari@uw.edu}}
\footnotetext[3]{\texttt{\{xiangyun, amir, dongki, shy, ali\}@fieldai.com}}
\endgroup

\thispagestyle{empty}
\pagestyle{empty}

\begin{abstract}
Robots must adapt to diverse human instructions and operate safely in unstructured, open-world environments. Recent Vision–Language models (VLMs) offer strong priors for grounding language and perception, but remain difficult to steer for navigation due to differences in action spaces and pretraining objectives that hamper transferability to robotics tasks. Towards addressing this, we introduce \ourmodel{}, a vision–language navigation system that finetunes internet-pretrained image diffusion models for path planning. Instead of directly predicting low-level actions, \ourmodel{} generates a path mask (i.e. a visual plan) in image space that captures fine-grained, context-aware navigation behaviors. A lightweight behavior-cloning policy grounds these visual plans into executable trajectories, yielding an interface that follows natural language instructions to generate diverse robot behaviors. To scale training, we supervise on path masks derived from self-supervised tracking models paired with VLM-augmented captions, avoiding manual pixel-level annotation or highly engineered data collection setups. In extensive real-world evaluations, \ourmodel{} outperforms state-of-the-art foundation model baselines on object reaching, obstacle avoidance, and terrain preference tasks, improving success rates by 33\% and reducing collisions by 54\% across both seen and unseen scenarios. Notably, we find that \ourmodel{} generalizes to unseen combinations of distinct tasks, revealing emergent compositional capabilities. Videos, code, and additional materials: \href{https://venturapath.github.io}{\textcolor{navy}{https://venturapath.github.io}}.
\end{abstract}


\section{Introduction}
\seclabel{introduction}

Mobile robots deployed in diverse, unstructured environments have untapped potential in domains such as construction inspection~\cite{rouvcek2019darpa}, urban maintenance~\cite{parker1998robotics}, and last-mile delivery~\cite{zhang2025crestescalablemaplessnavigation}. In these settings, robots must adapt their behavior to changing human preferences and environmental contexts. For example, at a construction site, a robot should avoid areas marked by caution tape, but it may enter if instructed by a worker to perform an inspection. In residential neighborhoods, robots should generally avoid disturbing private lawns, but may cut across to take out the trash when directed by a homeowner. Because situations often change quickly and unpredictably, robots must be able to adapt their behaviors rapidly based on diverse human instructions.

Language is a natural interface for conveying human intent, making it a flexible tool for adaptive autonomy in the open world. Recently, Vision-Language-Action (VLA) models output actions that enable robots to follow language instructions~\cite{brohan2023rt, zitkovich2023rt, black2024pi, sathyamoorthy2024convoi, vlmap}. By leveraging internet-scale data, VLAs have shown promising open-set image and language understanding capabilities~\cite{black2024pi, nasirianypivot}. However, existing VLA navigation systems struggle to ground language instructions in \emph{precise} robot motions. For example, methods \cite{vlmap,chang2024goat,hirose2025lelan,shah2023lm} based on CLIP-style \cite{radford2021learning} embedding typically use language only to locate the target (e.g. \emph{go to the red chair}) due to their limited language-conditioned planning capabilities. Systems that use transformer-based VLMs can plan at a coarse level using pre-defined image markers \cite{sathyamoorthy2024convoi} or discrete actions \cite{navid}. However, consider the instruction ``\emph{Keep a safe distance from kids}'': a robot must not only understand the meanings of ``safe'' and ``kids'', but also generate precise motions to satisfy the intent. As of today, it remains an open question how to most effectively leverage the open-world knowledge in foundation models and ground it in precise navigation plans.

Driven by this question, we propose a new architecture \ourmodel{}, that leverages pretrained image diffusion models~\cite{rombach2022high} for planning. Like denoising an image from language description, \ourmodel{} denoises a path mask (e.g. a ``visual plan'', see ~\figref{mainapproach}) to represent the robot's intended path across the scene. By formulating planning as an image generation problem, \ourmodel{} leverages the rich visual-linguistic priors and strong image generation ability of diffusion models to render realistic and instruction-aligned visual plans. A lightweight Behavior-Cloning (BC) policy is sufficient to convert the visual plans into executable waypoints, thereby enabling \ourmodel{} to ground diverse language commands in precise actions.

Notably, \ourmodel{} uses a visual tracking approach to automatically construct the ground truth path masks. In this way, we obtain pixel-precise and natural-looking path masks that natively handle occlusions. It does not need robot odometry, an assumption required by current approaches~\cite{hirose2025lelan, shah2023gnm, sridhar2024nomad}. Akin to the classifier-free guidance training in image-diffusion models, we train \ourmodel{} with a mix of 8.5 hours unlabeled robot videos and 1.5 hours of language-trajectory data. This makes \ourmodel{} potentially scalable to millions of internet videos. Combined, the high-quality groundtruths and label-efficient training scheme bolster \ourmodel{}'s generalization capabilities and precision. 

We evaluate \ourmodel{} in challenging outdoor environments, finding that \ourmodel{} outperforms SOTA VLA navigation systems on a variety of common navigation tasks ranging from terrain-aware navigation, obstacle avoidance, and object-centric goal reaching. Our contributions are as follows: 1) A simple finetuning protocol that adapts image diffusion models for multi-task path planning, 2) A scalable label generation pipeline for image space planning from unstructured robot demonstrations, and 3) An open-source language-captioned navigation dataset to support future research towards VLA models for navigation.

\begin{figure*}[htbp]
\begin{center}
    \centering
    \vspace{5pt}
    \includegraphics[width=1.0\textwidth]{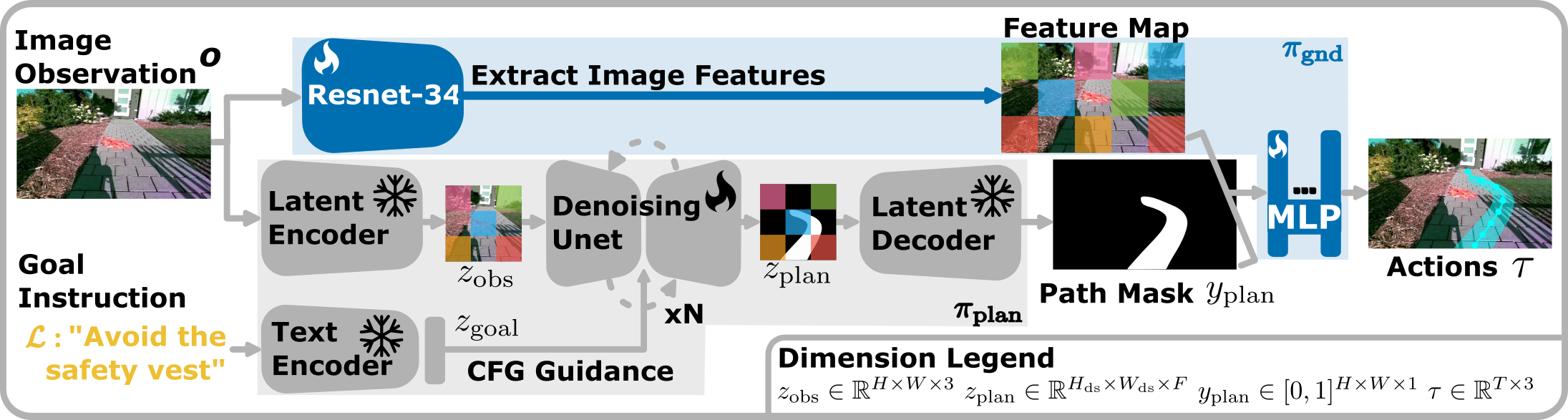}
    \caption{\ourmodel{} Architecture Overview. Our model is composed from an image diffusion planner $\pi_\text{plan}$ and grounding policy $\pi_\text{gnd}$. $\pi_\text{plan}$ generates a ``visual plan" in the form of a path mask, guiding $\pi_\text{gnd}$ to generate a sequence of xyz waypoints that satisfy the goal instruction $\mathcal{L}$.}
    \figlabel{architectureoverview}
    \vspace{-20pt}
\end{center}%
\end{figure*}

\section{Related Work}
\seclabel{related-work}

\textbf{Learning-based Navigation.} Driven by demands for models that understand diverse goal instructions and intricate affordances, recent works have shifted towards learning-based methods for robot navigation. These approaches range from general-purpose, single-task, and multi-task navigation models. General-purpose models~\cite{zhang2025crestescalablemaplessnavigation, sridhar2024nomad, roth2024viplanner} learn task-agnostic policies that reason about various environmental factors for producing safe paths. Single-task models~\cite{mao2025pacer, schmittle2025long, song2025vl, chang2024goat} learn specialized costs or actions to achieve a single objective, but must be retrained for each new task. Multi-task models~\cite{sathyamoorthy2024convoi, shahvint} seek to unify these works, learning a policy capable of following multiple tasks and constraints. Achieving this requires models that generalize across a combinatorial set of tasks and environments, and is typically achieved by leveraging large pre-trained foundation models~\cite{simeoni2025dinov3, beyer2024paligemma}. While these models offer internet-scale priors that make this problem tractable, adapting them for robotics tasks requires overcoming novel challenges discussed in the next section.

\textbf{Adapting Pre-trained Vision-Language Models.} With the emergence of vision–language models (VLMs) trained on internet-scale datasets, several works have explored adapting them for navigation. These methods typically rely on prompting VLMs for tasks that resemble their pre-training objectives, such as annotating images~\cite{nasirianypivot}, selecting between in-context examples~\cite{sathyamoorthy2024convoi, shah2023lm}, or performing visual question answering (VQA)~\cite{navid} to ground language instructions to robot actions. Other efforts fine-tune VLMs into vision–language–action (VLA) models to directly produce robot actions~\cite{cheng2024navila, black2024pi}, promoting more precise control. However, due to the significant differences between the original pre-training tasks and output space, these approaches struggle to follow semantically diverse task instructions and generate myopic local plans to reach long-horizon goals.

\textbf{Learning from Robot Foundation Models and Internet Data.} A complementary line of work directly distills affordance priors and actions from robot foundation models using large collections of internet data. These approaches~\cite{hirose2025lelan, mao2025pacer} condition on natural language or preference instructions to regress actions generated by an oracle navigation policy. While effective for following simple commands (e.g. \emph{go to object x}), these methods struggle to accommodate multiple tasks or generalize beyond the training data. Moreover, methods that directly predict robot actions require supervision from robot odometry, limiting their scalability in domains where accurate odometry data is difficult to obtain, such as internet videos.

\textbf{Relation to Prior Work.} Transformer-based VLMs, while understanding high-level visual semantics, struggle with fine-grained spatial reasoning and planning \cite{yang2025thinking}. In comparison, image diffusion models can generate high-fidelity images that align with language descriptions precisely. By formulating the navigation planning problem as an image generation problem, diffusion models can be more effective \emph{visual planners}.

In terms of goal-conditioning, our work is most similar to LeLAN~\cite{hirose2025lelan}, which conditions on object-goal language instructions. While LeLAN is limited to single-task conditioning, \ourmodel{} generalizes to diverse language instructions, enabling a multi-task policy that better aligns with open-world navigation demands. Methodologically, our approach also differs from prior efforts that employ diffusion models for navigation. Image-goal-conditioned policies~\cite{shahvint} use diffusion models to generate intermediate image subgoals for exploration, while diffusion policies~\cite{liang2024dtg, sridhar2024nomad} directly synthesize robot actions. By contrast, \ourmodel{} leverages internet-scale priors from Stable Diffusion~\cite{rombach2022high} to plan full trajectories directly in image space before grounding them into the robot’s action space, providing a structured and interpretable representation that supports diverse, language-conditioned tasks.

\section{Overview}
\seclabel{overview}

Our approach addresses the task-conditioned path planning problem, where the robot receives local camera observations $o \in \mathcal{O}$ and a language-specified task instruction $\mathcal{L}$, and must plan a path $\Gamma$ to accomplish the task~\cite{spaan2012partially}. Similar to prior work on multi-task robot learning~\cite{barreiros2025careful}, we identify two tasks as distinct if they differ in what they optimize for and their task-specific constraints. For instance, object goal navigation requires precise maneuvering to an object whereas abiding by terrain preferences requires maneuvering on the most desirable terrain available.

To accomplish the task, the robot is given a policy $\pi: (o, \mathcal{L}) \rightarrow \mathcal{A}$ mapping the current observation and task tuple to a sequence of primitive actions $a \in \mathcal{A}$, where $a\in \mathbb{R}^3$ represents a sequence of xyz Cartesian waypoints. We assume the robot continuously replans given new observations, tracing a path $\Gamma^\pi$. The robot's objective is to generate a path $\Gamma^\pi$ that lies within the set of acceptable paths given by the expert policy such that $\Gamma^\pi \in \{ \Gamma  |  \Gamma \sim \pi^E \}$.

\begin{figure*}[htbp]
\begin{center}
    \centering
    \vspace{5pt}
    \includegraphics[width=\textwidth]{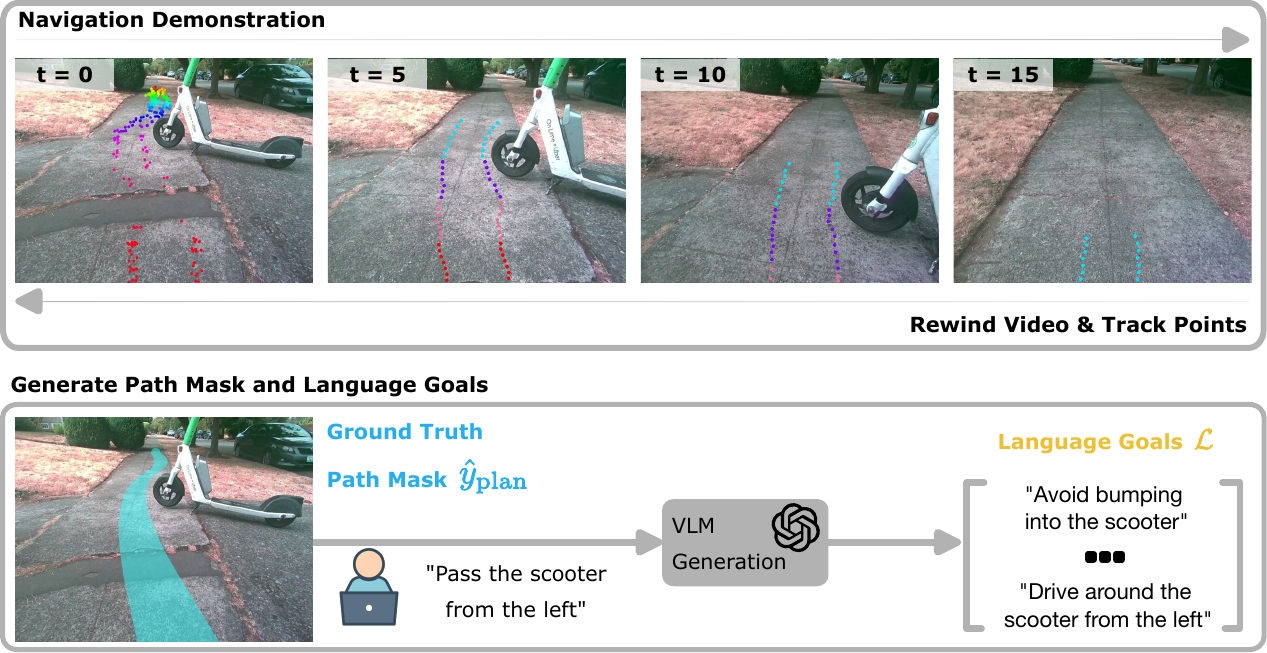}
    \caption{\ourmodel{} Data Generation Pipeline. We generate ground truth path masks by using an off-the-shelf tracker~\cite{karaev2024cotracker3} to track points traversed by the robot in reversed videos. Each dots' color corresponds to the first time that it is tracked from. We provide seed captions to a VLM to generate diverse language goals that explain the path.}
    \figlabel{datapipeline}
    \vspace{-20pt}
\end{center}%
\end{figure*}
\section{Approach}
\seclabel{approach}

We posit that adapting an image diffusion model to generate image-space plans is a highly effective way to transfer internet-scale semantic knowledge into navigation policies. To make this scalable, we exploit advances in off-the-shelf point tracking~\cite{karaev2024cotracker3} to automatically extract plan labels from uncalibrated egocentric video, enabling supervision from diverse, unstructured data. Building on these ideas, \ourmodel{} introduces two main components: a diffusion-based planner $\pi_\text{plan}$ (see~\figref{architectureoverview}) that performs task-conditioned path planning in image space, and an auto-labeling pipeline (see~\figref{datapipeline}) that provides the supervision needed to train $\pi_\text{plan}$. In the remainder of this section, we describe the \ourmodel{} architecture, training objective, and scalable auto-labeling pipeline.

\subsection{\ourmodel{} Architecture}
\sseclabel{approach:architecture}

Our architecture is composed of two components, a language-conditioned image diffusion policy $\pi_\text{plan}$ that generates path masks, and a grounding policy $\pi_\text{gnd}$ that grounds these visual plans to trajectory waypoints (see~\figref{architectureoverview}).

We initialize $\pi_\text{plan}$ from a pre-trained text-to-image latent diffusion model (Stable Diffusion v2~\cite{rombach2022high}) and freeze the variational autoencoder (VAE) and text encoder for the duration of training. We unfreeze the latent diffusion U-Net so that $\pi_\text{plan}$ can adapt its denoising process for path mask generation. Our planner $\pi_\text{plan}$ encodes the image observation $o$ and natural language goal instruction $\mathcal{L}$ using the pre-trained image and text encoders to obtain a latent image $z_\text{rgb}$ and goal $z_\text{goal}$. We sample an image $\hat{z}_\text{mask}$ from Standard noise and stack $\hat{z}_\text{mask}$ and $z_\text{rgb}$ along the channel dimension. Our latent diffusion U-Net conditions on a stacked feature map consisting of the latent image and goal $z_\text{goal}$, and learns to denoise a latent path mask $z_\text{plan}$. Finally, we decode $z_\text{plan}$ using the frozen VAE decoder. Since the pre-trained VAE decodes three-channel images, we average along the channel dimension to obtain a scalar likelihood map for the final image space plan $y_\text{plan}$.

\ourmodel{} implements the grounding policy $\pi_\text{gnd}$ using a ResNet-34~\cite{he2016deep} to encode the current observation $z_\text{obs}$ and stacks $z_\text{obs}$ with $y_\text{plan}$ along the channel dimension to construct the context vector $c$. We pass $c$ to a Spatial Convolution~\cite{finn2016deep} layer before using a Multilayer Perceptron (MLP) to predict a sequence of xyz waypoint targets.

\subsection{\ourmodel{} Objective}
\sseclabel{approach:objective}

The planner $\pi_\text{plan}$ and grounding policy $\pi_\text{gnd}$ are trained in two stages with the following loss function:
\begin{equation}
    \mathcal{L}_\text{VENTURA} = \mathcal{L}_\text{plan} + \mathcal{L}_\text{gnd}.
\end{equation}
For diffusion training, we approximate the conditional distribution $p(y_\text{plan} | o, \mathcal{L})$, where $\mathcal{L}$ is the language task instruction. In the \textit{forward} process, we start from $y_{\text{plan},0}:=y_\text{plan}$ and gradually add Gaussian noise at levels $t\in \{1,...,T\}$ to obtain noisy samples $y_{\text{plan},t}$:
\begin{equation}
    y_{\text{plan},t} = \sqrt{\alpha_t} y_{\text{plan},0} + \sqrt{1-\alpha_t} \epsilon,
\end{equation}
where $\epsilon \sim \mathcal{N}(0, I)$, $\alpha_t=\prod^t_s=1 - \beta_s$, and $\{ \beta_1,...,\beta_T\}$ is the process variance schedule. We train $\pi_\text{plan}$ to predict the image plan by gradually removing noise in the \textit{reverse process}. 

At train time, we sample a data point $(o, y_\text{plan}, \mathcal{L})$ and inject  $\epsilon$ noise from a random timestep $t$ to obtain the noise estimate $\hat\epsilon = \epsilon(y_\text{plan, t}, o, \mathcal{L},t)$ and minimize the standard diffusion objective~\cite{ho2020denoising}:
\begin{equation}
\mathcal{L}_\text{plan} = \mathbb{E}_{y_{\text{plan},0}, \epsilon~\sim\mathcal{N}(0, I), t\sim\mathcal{U}(T)}||\epsilon - \hat{\epsilon}||^2_2.
\end{equation}
At inference time, we iteratively apply $\pi_\text{plan}$ starting from  noise $y_{\text{plan},T}$ to reconstruct the true image space plan $y_{\text{plan},0}$. 

After training $\pi_\text{plan}$ to convergence, we freeze $\pi_\text{plan}$ and train the grounding policy $\pi_\text{gnd}$ to minimize the mean squared error (MSE) loss between the predicted and ground truth actions $a$ using the predicted image plan $y_\text{plan}$:
\begin{equation}
    \mathcal{L}_\text{gnd} = MSE(\pi_\text{gnd}(c) - a),
\end{equation}
where $c = \{y_\text{plan}, z_\text{obs}\}$. 

\subsection{Autolabeling Pipeline}
\sseclabel{approach:labels}

Previously, we described the \ourmodel{} objective, which assumes access to a ground truth path mask $\hat{y}_\text{plan}$ and robot actions $a$. In this subsection, we expand on how to automatically extract these ground truth masks.

Vision-based navigation models scale favorably with dataset size and diversity, motivating the need for flexible auto-labeling methods that work reliably across a variety of data collection setups. While it is possible to compute the robot's 3D position with a calibrated and synchronized hardware setup, this approach is unreliable for long trajectories, as small state estimation errors can cause points in lethal regions to be considered traversable. We address these limitations by adopting an approach described in LRN~\cite{schmittle2025off} that uses Co-Tracker~\cite{karaev2024cotracker3} to track masks in the image that correspond to the robot's future positions. Our approach generates masks that accurately represent the robot's path in the image without relying on accurate calibrations, human labels, or complex hardware setups.

To compute these masks, we play the video in reverse and drop a set of ``breadcrumb tracks" on the pixels beneath the robot at the bottom of the image. We track these points across the entire video sequence, adding new points every 0.25 seconds. For each frame, we use the visibility value predicted by Co-Tracker to determine the set of visible points and construct a binary segmentation mask that best fits these points.

\begin{figure*}[htbp]
\begin{center}
    \centering
    \vspace{6pt}
    \includegraphics[width=\textwidth]{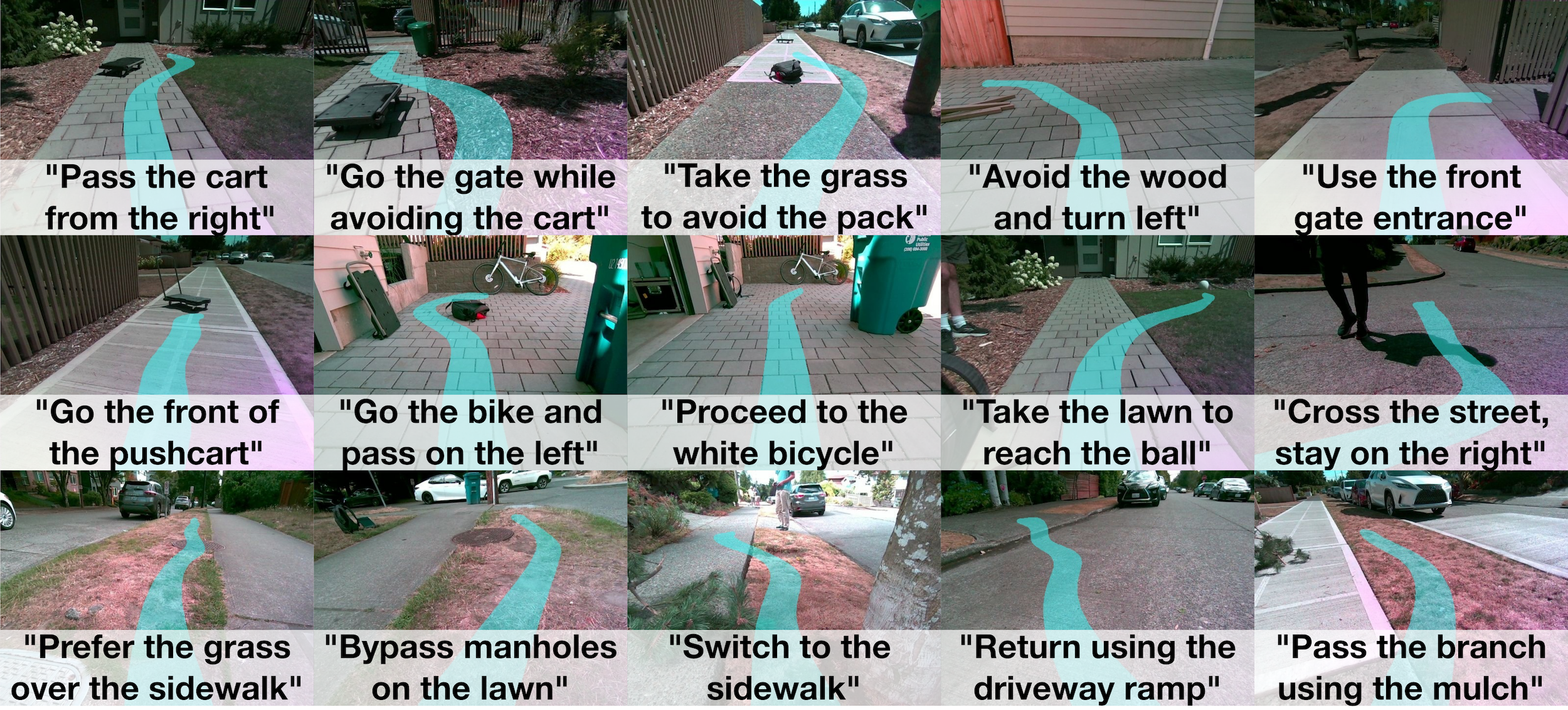}
    \caption{Ground truth language and path mask labels from the \ourmodel{} dataset. We co-train \ourmodel{} on a collection of navigation demonstrations with and without language captions. To bolster generalization to novel language prompts, we augment human-labeled captions using a pre-trained VLM to automatically generate diverse caption variations.}
    \figlabel{datasetexamples}
    \vspace{-20pt}
\end{center}%
\end{figure*}
\section{Implementation Details}
\seclabel{implementation}

In this section, we describe the experimental setup and model-specific details to ensure fair evaluation. All baselines are evaluated on a wheeled quadrupedal robot (Unitree GO2-W) using monocular RGB observations from an Intel Realsense RGB-D camera (depth not used). Each method predicts a sequence of 8 cartesian waypoints spaced 0.4m apart that are tracked using the same model predictive controller (MPC). 

To train \ourmodel{}, we collect a dataset of 10 hours of navigation demonstrations, which we describe in detail in~\secref{dataset}. We train the image planner $\pi_\text{plan}$ for 50 epochs with a learning rate of 3e-4 and a batch size of 512, adopting the same training settings as Marigold~\cite{ke2025marigold} for the remaining hyperparameters. Additionally, we train with a classifier-free guidance weight of 0.05 to learn a task-conditioned and task-agnostic planner. We train the grounding policy $\pi_\text{gnd}$ for another 50 epochs using the same settings as prior language-conditioned behavior cloning work~\cite{hirose2025lelan}.

\textbf{Baselines.} We evaluate \ourmodel{} against LeLaN~\cite{hirose2025lelan} and Convoi~\cite{sathyamoorthy2024convoi}, two SOTA robot foundation model and VLM methods that predict waypoint actions given RGB observations and language commands. We pre-train LeLAN on the same GNM~\cite{shah2023gnm} and Youtube tour dataset used in the original work for 100 epochs before finetuning on the same dataset split used by \ourmodel{} for another 100 epochs. We reproduce Convoi~\cite{sathyamoorthy2024convoi} as faithfully as possible since there is no open-source code release, removing the initial point cloud filtering safety layer to maintain fairness across each baseline.

\section{Dataset Details}
\seclabel{dataset}

The \ourmodel{} dataset consists of approximately 10 hours of navigation demonstrations, consisting of 8.5 hours of task-agnostic demonstrations and 1.5 hours of task-conditioned demonstrations. The task-agnostic demonstrations do not contain any unsafe actions, such as colliding into objects, and simply perform navigation to long-horizon goals. Our task-conditioned demonstrations are paired with language captions that describe behaviors such as going to objects, following spatial directions, following different terrain preferences, and avoiding objects. A small subset of these language captions and path masks are shown in~\figref{datasetexamples}. To generate corresponding language captions, a human labeler provides a short description to explain the observed navigation behavior. Then, we prompt gpt4o-mini~\cite{achiam2023gpt} with a short system prompt, annotated image, and human-generated caption to automatically generate diverse, semantically identical captions for training.

\begin{table}[t]
\begin{tabular}{|l|l|l|l|l|l|l|}
\hline
\textbf{Model}
  & \multicolumn{2}{l|}{\textbf{Obs. Avoidance} $\uparrow$ }
  & \multicolumn{2}{l|}{\textbf{Obj. Goal} $\uparrow$}
  & \multicolumn{2}{l|}{\textbf{Ter. Aware} $\uparrow$} \\ \hline
\textbf{}
  & \mc{\makecell{Seen}}
  & \mc{\makecell{Uns.}}
  & \mc{\makecell{Seen}}
  & \mc{\makecell{Uns.}}
  & \mc{\makecell{Seen}}
  & \mc{\makecell{Uns.}} \\ \hline
\ourmodel{}
  & \mc{\textbf{13/15}} & \mc{\textbf{4/5}}
  & \mc{\textbf{9/10}}  & \mc{\textbf{7/10}}
  & \mc{\textbf{6/6}}   & \mc{\textbf{5/6}} \\ \hline
\ourmodel{}-P
  & \mc{10/15} & \mc{1/5}
  & \mc{5/10} & \mc{4/10}
  & \mc{4/6} & \mc{2/6} \\ \hline
LeLaN~\cite{hirose2025lelan}
  & \mc{9/15} & \mc{1/5}
  & \mc{8/10} & \mc{3/10}
  & \mc{3/6}  & \mc{2/6} \\ \hline
Convoi~\cite{sathyamoorthy2024convoi}
  & \mc{8/15} & \mc{3/5}
  & \mc{7/10} & \mc{\textbf{7/10}}
  & \mc{4/6} & \mc{3/6} \\ \hline
\end{tabular}
\caption{\textbf{Multi-task planning evaluations.} \ourmodel{} consistently outperforms baselines across representative robot navigation tasks in seen and unseen environments. We define the success rate criteria for each task in~\secref{evaluation}. Bolded numbers indicate the best performing method(s) for each category. We use the following abbreviations: Obs. - Obstacle, Obj. - Object, Ter. - Terrain, Uns. - Unseen, \ourmodel{}-P - our approach without internet pre-training.}
\label{tab:multitask_evals}
\vspace{-10pt}
\end{table}

\begin{table}[t]
\begin{tabular}{|l|l|l|l|l|l|l|}
\hline
\textbf{Model}
  & \multicolumn{2}{l|}{\textbf{Short }$\uparrow$}
  & \multicolumn{2}{l|}{\textbf{Medium }$\uparrow$}
  & \multicolumn{2}{l|}{\textbf{Long }$\uparrow$} \\ \hline
\textbf{}
  & \mc{\makecell{Seen}}
  & \mc{\makecell{Uns.}}
  & \mc{\makecell{Seen}}
  & \mc{\makecell{Uns.}}
  & \mc{\makecell{Seen}}
  & \mc{\makecell{Uns.}} \\ \hline
\ourmodel{}
  & \mc{\textbf{6/6}} & \mc{\textbf{6/6}}
  & \mc{\textbf{6/6}} & \mc{\textbf{5/6}}
  & \mc{\textbf{5/6}} & \mc{\textbf{4/6}} \\ \hline
\ourmodel{}-P
  & \mc{3/6} & \mc{1/6}
  & \mc{3/6} & \mc{2/6}
  & \mc{3/6} & \mc{1/6} \\ \hline
LeLaN~\cite{hirose2025lelan}
  & \mc{6/6} & \mc{4/6}
  & \mc{4/6} & \mc{2/6}
  & \mc{2/6} & \mc{1/6} \\ \hline
Convoi~\cite{sathyamoorthy2024convoi}
  & \mc{\textbf{6/6}} & \mc{5/6}
  & \mc{5/6} & \mc{2/6}
  & \mc{3/6} & \mc{3/6} \\ \hline
\end{tabular}
\caption{\textbf{Long range planning evaluations.} We show the success rate of reaching object goal targets across 140 trials with 10 objects. A trial is deemed successful if the robot reaches with 0.5 m of the target object. Bolded numbers indicate the highest performing method(s) per category. We use the following abbreviations: Uns. - Unseen, \ourmodel{}-P - our approach without internet pretraining. }
\vspace{-10pt}
\label{tab:longrange_evals}
\end{table}

\begin{table}[t]
\centering
\begin{tabular}{|l|c|c|}
\hline
\textbf{Model} & \textbf{Mean $L_2$ Error $\downarrow$} & \textbf{Hausdorff Distance $\downarrow$} \\ \hline
\ourmodel{}                 & \textbf{0.04} & \textbf{0.08} \\ \hline
\ourmodel{}-P               & 0.06 & 0.09 \\ \hline
LeLaN~\cite{hirose2025lelan} & 0.09 & 0.17 \\ \hline
\end{tabular}
\caption{\textbf{Trajectory Error on the Test Set.} We compare the test set performance of each baseline that is trained on the \ourmodel{} dataset. Our approach achieves the lowest average $L_2$ error and Hausdorff distance, indicating that our method is able to generate more precise actions that closely match the expert behavior. Bolded numbers indicate the best performing method(s) for each metric.}
\tablabel{quantitative_evals}
\vspace{-15pt}
\end{table}
\section{Evaluation}
\seclabel{evaluation}
\begin{figure*}[htbp]
\begin{center}
    \centering
    \vspace{6pt}
    \includegraphics[width=\textwidth]{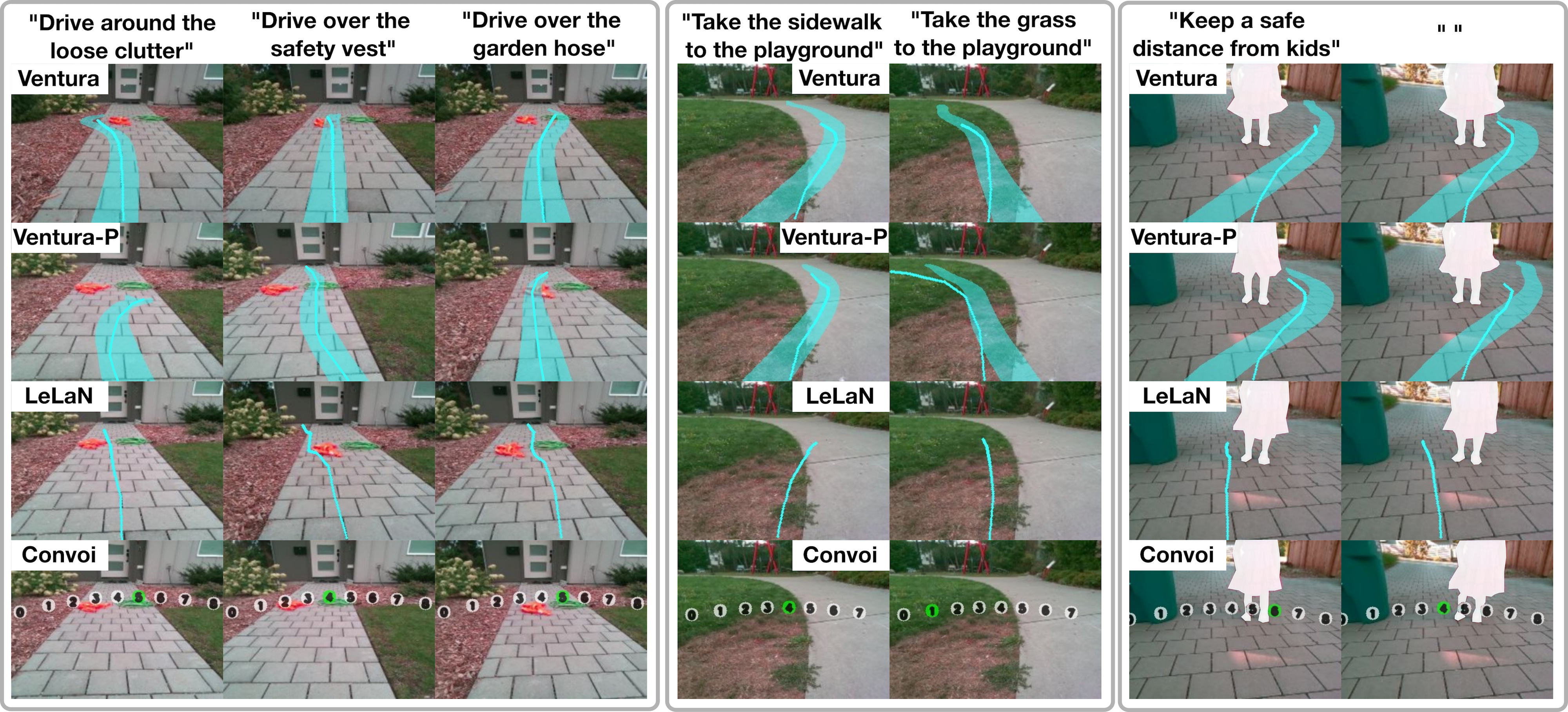}
    \caption{\textbf{Qualitative analysis of various visual-language navigation baselines.} \ourmodel{} consistently outperforms existing approaches in terms of task alignment and generalization to unseen entities and environments. We use the following abbreviations: \ourmodel{}-P: Our model without initializing with internet-pretrained weights.}
    \figlabel{qualitativeresults}
    \vspace{-20pt}
\end{center}%
\end{figure*}

We evaluate \ourmodel{} in 2 seen and 2 unseen outdoor environments and answer the following questions to understand the importance of our contributions and overall performance on multi-task and task-agnostic navigation.
\begin{itemize}[]
    \item (\textit{$Q_1$}) Does \ourmodel{} improve success rate on diverse navigation tasks compared to SOTA approaches that leverage pre-trained foundation models?
    \item (\textit{$Q_2$}) Is \ourmodel{} able to use semantic knowledge from pre-trained foundation models to improve generalization performance?
    \item (\textit{$Q_3$}) Does \ourmodel{} improve the success rate on tasks that require long-range planning?
\end{itemize}

To investigate the preceding questions, we conduct more than 150 obstacle avoidance, object goal navigation, and preference-aware terrain navigation experiments against LeLaN~\cite{hirose2025lelan} and Convoi~\cite{sathyamoorthy2024convoi}. Our test environments feature a diverse set of objects and terrains ranging from common entities like trash cans and sidewalks to rare entities like safety vests and playgrounds. We evaluate each method using success rate as the primary criteria, classifying trials as failures if the robot does not reach within 0.5m of the goal, collides with an obstacle, or drives on unfavorable terrain for more than 2 seconds. 

Towards understanding $\mathcal{Q}_1$, we observe in~\tabref{multitask_evals} that \ourmodel{} outperforms all other approaches in seen and unseen environments by 40\% and 33\% respectively on average across all tasks. This is consistent with the results in~\tabref{quantitative_evals}, demonstrating that our method is able to plan paths that align more closely with expert behavior. Specifically, we find that our approach is able to ground diverse actions to unseen entities even under instruction ambiguity. We highlight this behavior in~\figref{qualitativeresults}, where \ourmodel{} correctly avoids an unseen safety vest and garden hose when told to ``avoid loose clutter". Furthermore, the same model can rapidly adapt its behavior to align with more specific instructions, such as ``drive over the garden hose". By comparison, while LeLAN and Convoi can follow specific instructions, they struggle to infer user intent when given ambiguous commands like "avoid loose clutter". We also observe that \ourmodel{} is able to generate precise motion commands that respect nuanced commands like ``keep a safe distance from kids". From these results, we conclude that \ourmodel{} is significantly more effective at interpreting language commands and identifying collision-free paths across common navigation tasks.

Towards \textit{$Q_2$}, we compare \ourmodel{} with and without StableDiffusion weight initialization to understand how internet pre-training on non-robotics tasks transfers to robot path planning. We observe that initializing the denoising Unet with StableDiffusion improves overall performance by 47\% and 128\% on average across seen and unseen scenarios respectively compared to training from scratch (\ourmodel{}-P).~\figref{qualitativeresults} corroborates these findings, showing that the model trained from scratch struggles to identify unseen entities, often behaving randomly for object-centric goal navigation when presented with multiple unseen options. Even with these limitations, \ourmodel{}-P performs on par with LeLaN despite being trained on far less robot data. We hypothesize that this is possible because the StableDiffusion text and image encoders are pre-trained on more diverse data sources than those used by robot foundation models, enabling better zero-shot generalization compared to learning these model components from scratch.

To understand \textit{$Q_3$}, we vary the target object distance for the object goal navigation task (4m, 8m, 12m) and compare each model's ability to perceive and plan towards long-range entities. From~\tabref{longrange_evals}, we find that while existing approaches perform comparably to \ourmodel{} in short to medium ranges, our approach separates itself for longer distances, outperforming the second best approach, Convoi, by 21.4\% and 50.0\% on average across seen and unseen scenarios respectively. Interestingly, we observe that the most common sources of failures are caused by an inability to localize the target object and losing sight of the target object. As seen in~\figref{qualitativeresults}, LeLaN and Convoi plan paths that close the target distance, but neglect to consider how the future path affects the visibility of the target object. This introduces failures where the target object gradually drifts out of the field of view. In contrast, \ourmodel{} predicts path masks directly in the image, resulting in precise, long range plans that reduce the likelihood of these kinds of myopic decisions.

\begin{figure}
    \centering
    \vspace{5pt}
    \includegraphics[width=\linewidth]{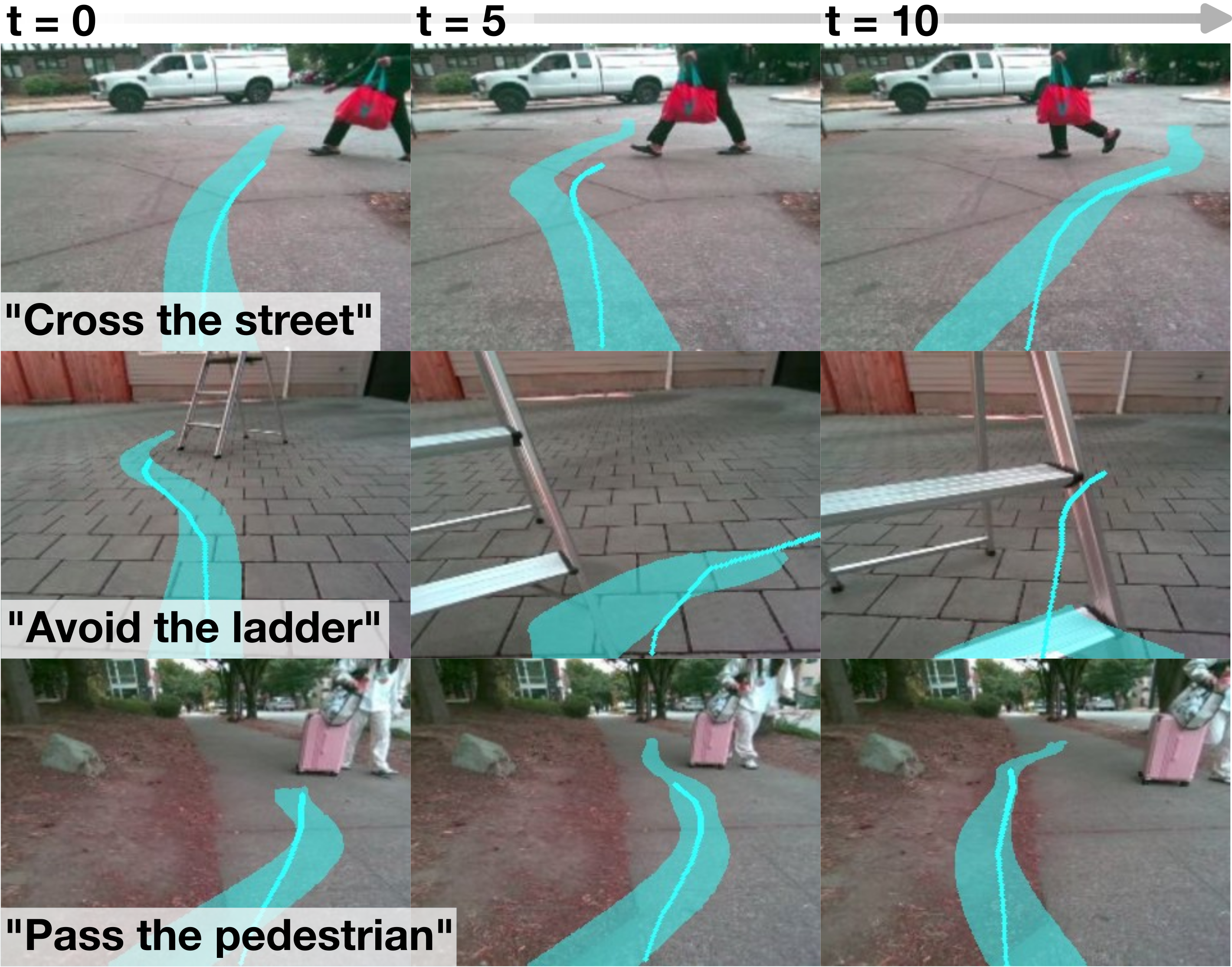}
    \caption{Limitations of \ourmodel{}. In the first row, our approach struggles to infer social dynamics, leading to path plans that cut in front of the pedestrian. In the second row, we test our model's ability to handle tight turns when avoiding obstacles. It is difficult to represent backwards actions as visual plans, leading to suboptimal behavior when backtracking is the only valid path. In the final row, we observe that while \ourmodel{} is able to pass the pedestrian safely, the generated plans are often temporally inconsistent, occasionally resulting in unstable behavior.}
    \figlabel{limitations}
    \vspace{-15pt}
\end{figure}
\section{Limitations and Future Work}
\seclabel{limitations}

While \ourmodel{} inherits open-set semantic knowledge from pre-trained foundation models, it does not enable generalization to novel motion primitives. This limits our model's ability to follow complex motion patterns not seen in the training data, such as ``circling around the house". Furthermore, it is difficult to capture motion dynamics with visual plans, which are important for scenarios depicted in~\figref{limitations} with dynamic agents (e.g. social navigation) or complex vehicle dynamics (e.g. offroad driving). Another promising direction to explore is extending \ourmodel{} to reason about multiple observations and produce temporally consistent plans. This will enhance robustness in long horizon partially observable environments that require joint understanding of information from multiple viewpoints. 

\section{Conclusion}
\seclabel{conclusion}

In this paper, we presented \ourmodel{}, a flexible vision-language model that repurposes pre-trained image diffusion models to plan paths that follow diverse language instructions. Our unified policy uses a pre-trained image diffusion backbone pre-trained for image generation to generate path masks (i.e. visual plans) conditioned on language commands. We train a lightweight behavior cloning policy to ground these path masks to robot actions, demonstrating its robustness and generalizability to novel environments despite limited on-robot training data. We study our approach's effectiveness across a variety of navigation environments and tasks, showing improvements of up to 33\% in performance compared to SOTA in unseen settings. Based on these findings, we believe that \ourmodel{} presents a promising direction for leveraging internet-scale priors to achieve adaptive, open-world autonomy.

\section{Acknowledgements}
\seclabel{acknowledgements}
We thank Amy Zhang, Jay Patrikar, Sebastian Scherer, Fadhil Ginting, and Max Smith for their support, advice, comments, and discussions during the project. Furthermore, the authors would like to thank Abdullah Chaudhry, Andrew Shim, Tarran Vail, Harmish Khambhaita, Eugene Lo, and Yaseen Elhalafway for their assistance in maintaining and deploying the robots used during the duration of this project.

\printbibliography[]

\clearpage

\end{document}